\documentclass{nolta}

\usepackage[dvips]{graphicx}

\usepackage{url}

\usepackage{bm}

\usepackage{amsmath}

\usepackage{array,booktabs}

\usepackage{subcaption}

\Vol{13}%
\No{2}%
\Year{2022}%
\Month{4}%

\title{Differential evolution with perturbation 
for continuous function optimization\\ 
}

\AUTHOR{%
\author{Takuya Muto}{1}, 
\author[a]{Michiharu Maeda}{1}
}

\AFFILIATE{%
\affiliate{Department of Computer Science and Engineering, Fukuoka Institute of Technology\\
3-30-1 Wajirohigashi, Higashi-ku, Fukuoka, Fukuoka 811-0295, Japan}{1},
}

\EMAIL{%
\email{maeda@fit.ac.jp}{a}
}

\received{10}{15}{2021}
\revised{12}{18}{2021}
\published{4}{1}{2022}

\begin{document}

\begin{abstract} 
  This paper presents differential evolution with perturbation (DEP) for 
  continuous function optimization. Differential evolution is a simple 
  and efficient algorithm. Perturbation is the phenomenon that the motion 
  by the contribution of the main force is disturbed by the influence 
  subsidiary power. In order to show the effectiveness of the proposed 
  algorithm, we compare it to existing algorithms by CEC2017 datasets. 
\end{abstract}

\begin{keywords}
  differential evolution, perturbation, function optimization
\end{keywords}

\maketitle 

\section{Introduction}

Metaheuristics organically combine empirical approaches for 
solving optimization problems ~\cite{REF Metaheuristics}. 
In the real world, 
metaheuristics are used as algorithmic frameworks without depending on problems 
that provides series of guidelines or strategies. Metaheuristics require 
an approximately solution within an acceptable time. 
As the performance of metaheuristics varies depending on the parameter 
settings, metaheuristics are expected to have the potential to search 
better solutions. 
Metaheuristics include, real-coded genetic algorithm (RGA), 
particle swarm optimization (PSO), and differential evolution (DE). 
RGA adopts a real number vector as a coding approach and implements 
a crossover technique on it ~\cite{REF RGA}. PSO is an approach of searching for the 
optimum solution while sharing the information of the best solution 
using particles with velocity and position information ~\cite{REF PSO}. DE is an 
algorithm that randomly extracts a candidate solution from a set of 
candidate solutions and repeats mutation and crossover to generate 
a new solution ~\cite{REF DE}. 
DE has been applied to various practical problems 
due to its simplicity and efficiency. DE is applied in many fields. 
In the field of photovoltaics, a global search space differential 
evolution algorithm for tracking the global maximum power point (GMPP) 
has been proposed ~\cite{REF DE with MPPT}. To minimize the energy consumed by an active 
magnetic bearing (AMB), a recurrent wavelet fuzzy neural network 
with adaptive differential evolution based dynamic bias current control 
strategy is suggested ~\cite{REF DE with AMB}. An approach to improve the quality of service 
provided by the taxi system by optimizing the fuzzy controller through 
differential evolution has been studied ~\cite{REF DE with Taxi}. The integration of reinforcement 
learning-based differential evolution with the cooperative coevolution (R-CCDE) 
method in a compensatory neuro-fuzzy controller (CNFC) is presented ~\cite{REF DE with CNFC}. 
On the other hand, DE is difficult to use for certain types of functions. Besides, even if the population does not converge to a 
local solution, the search is stagnant because the algorithm cannot find a better solution ~\cite{REF EDE}. 
There exist DE variants, such as JADE ~\cite{REF JADE} and SHADE ~\cite{REF SHADE}. 
JADE is an approach that automatically adjusts the parameters $C_{r}$ and $F$ 
and speeds up the convergence of the solution using a novel mutation strategy called current-to-$p$best/1. SHADE is a modified 
approach of JADE which uses different parameter adaptation mechanism.

In this paper, we present differential evolution with perturbation (DEP). 
Perturbation is a physical phenomenon in which a main force is 
disturbed by a subsidiary power. 
By introducing the perturbation, our algorithm searches for a  solution within a certain range. 
Therefore, the algorithm with perturbation tends to be able to explore a solution efficiently and to avoid to local optima. 
In order to show the effectiveness 
of the proposed algorithm, we compare it to existing algorithms by 
CEC2017 datasets. 

\section{Differential Evolution}

Differential evolution (DE) is a simple and efficient algorithm. DE 
generates a new solution by selection, mutation, and crossover. Selection 
is to extract three solutions randomly from a set of candidate solutions. 
Mutation is to generate a mutant solution using the three selected solutions. 
Mutation is expressed by the following formula. 
\begin{equation}
	\bm{v_{i,j}}=\bm{x_{a,j}}+F(\bm{x_{b,j}}-\bm{x_{c,j}})
\end{equation}
where $\bm{v}$ is a mutant solution, $i$ is the solution number, and $j$ is the 
dimension number. $\bm{x_{a}}$, $\bm{x_{b}}$, and $\bm{x_{c}}$ are selected 
solutions randomly from the candidate solutions. $F$ indicates the scale factor. 
Crossover is to generate a trial solution using the mutant solution and a target 
solution. DE crosses in the following formula. 
\begin{equation}
	\bm{u_{i,j}} = \begin{cases}
		\bm{v_{i,j}} & \text{if}(\mathit{rand_{j}}\leq{C_{r}} \text{ or }j=j_{\mathit{rand}})\\
		\bm{x_{i,j}} & \text{otherwise}
	\end{cases}
\end{equation}
where $\bm{u}$ is a trial solution, and $\bm{x}$ is a target solution. $\mathit{rand_{j}}$ 
is uniformly generated by random number in [0,1]. $C_{r}$ denotes the crossover rate. 
$j_{\mathit{rand}}$ is an index chosen randomly. The pseudo code of DE is 
shown in Algorithm 1. 

\begin{center}
	\begin{tabular}{l} \toprule
		{\bf Algorithm 1:} Differential Evolution \\ \midrule
		Initialize parameters: $N$, $D$, $MaxFEs$, $F$, and $C_{r}$. \\
		Generate randomly initial solutions $\bm{x_{i}} (i=1,2,3,...,N)$. \\
		Calculate the fitness of initial solutions. \\
		Set the best solution $\bm{x_{best}}$ according to the fitness and $FEs=N$. \\
		{\bf while} $FEs < MaxFEs$ \\
		\hspace{5mm}{\bf for} $i=1$ to $N$ \\
		\hspace{10mm}Select three solutions $\bm{x_{a}}$, $\bm{x_{b}}$, and $\bm{x_{c}}$ randomly from the candidate solutions. \\
		\hspace{10mm}Generate the mutant solution $\bm{v_{i}}$ according to (1). \\
		\hspace{10mm}Generate the trial solution $\bm{u_{i}}$ according to (2). \\
		\hspace{10mm}Calculate the fitness of the trial solution $\bm{u_{i}}$ and $FEs=FEs+1$. \\
		\hspace{10mm}{\bf if} the fitness of $\bm{u_{i}}$ is better than that of $\bm{x_{i}}$. \\
		\hspace{15mm}$\bm{x_{i}}=\bm{u_{i}}$. \\
		\hspace{15mm}{\bf if} the fitness of $\bm{x_{i}}$ is better than that of $\bm{x_{best}}$. \\
		\hspace{20mm}$\bm{x_{best}}=\bm{x_{i}}$. \\
		\hspace{15mm}{\bf end if} \\
		\hspace{10mm}{\bf end if} \\
		\hspace{5mm}{\bf end for} \\
		{\bf end while} \\
		Output the best solution $\bm{x_{best}}$ \\ \bottomrule
	\end{tabular}
\end{center}

\section{Differential Evolution with Perturbation} 

In this section, we propose an algorithm of differential evolution with 
perturbation (DEP). 
Perturbation is the phenomenon that the motion by the contribution of the 
main force is disturbed by the influence subsidiary power. The new solution 
becomes the better solution comparing the current solution with the perturbed 
solution. The perturbation is added when current function evaluation $FEs$ is less than 
one quarter of maximum number of function evaluation $MaxFEs$. 
The solution with perturbation is determined by the following equation. 
\begin{equation}
	\bm{u_{i,j}^{'}} = \begin{cases}
		\bm{v_{i,j}}+sI\bm{r_{j}} & \text{if}(\mathit{rand_{j}}\leq{C_{r}} \text{ or }j=j_{\mathit{rand}})\\
		\bm{x_{i,j}}+sI\bm{r_{j}} & \text{otherwise}
	\end{cases}
\end{equation}
where $\bm{u^{'}}$ indicates the solution with perturbation. $\bm{r}$ is a vector 
generated randomly and $||\bm{r}||\leq{1}$. $s$ and $I$ are calculated in the 
following formula. 
\begin{equation}
	s=a-4aFEs/MaxFEs
\end{equation}
\begin{equation}
	I=(x_{max}-x_{min})/R
\end{equation}
where $a$ and $R$ are constant parameters. $x_{max}$ and $x_{min}$ are maximum 
and minimum in the search range. The pseudo code of DEP is shown in Algorithm 2. 

\begin{center}
	\begin{tabular}{l} \toprule
		{\bf Algorithm 2:} Differential Evolution with Perturbation \\ \midrule
		Initialize parameters: $N$, $D$, $MaxFEs$, $F$, and $C_{r}$. \\
		Generate randomly initial solutions $\bm{x_{i}} (i=1,2,3,...,N)$. \\
		Calculate the fitness of initial solutions. \\
		Set the best solution $\bm{x_{best}}$ according to the fitness. \\
		$FEs=N$. \\
		{\bf while} $FEs < MaxFEs$ \\
		\hspace{5mm}{\bf for} $i=1$ to $N$ \\
		\hspace{10mm}Select three solutions $\bm{x_{a}}$, $\bm{x_{b}}$, and $\bm{x_{c}}$ randomly from the candidate solutions. \\
		\hspace{10mm}Generate the mutant solution $\bm{v_{i}}$ according to (1). \\
		\hspace{10mm}{\bf if} $FEs<MaxFEs/4$ \\
		\hspace{15mm}Calculate the trial solution $\bm{u_{i}}$ according to (2). \\
		\hspace{15mm}Calculate the fitness of trial solution $\bm{u_{i}}$ and $FEs=FEs+1$. \\
		\hspace{15mm}Calculate the trial solution with perturbation $\bm{u_{i}^{'}}$ according to (3). \\
		\hspace{15mm}Calculate the trial solution with perturbation $\bm{u_{i}^{'}}$ and $FEs=FEs+1$. \\
		\hspace{15mm}{\bf if} The fitness of $\bm{u_{i}}$ is better than that of $\bm{u_{i}^{'}}$. \\
		\hspace{20mm}$\bm{u_{i}}=\bm{u_{i}}$. \\
		\hspace{15mm}{\bf else} \\
		\hspace{20mm}$\bm{u_{i}}=\bm{u_{i}^{'}}$. \\
		\hspace{15mm}{\bf end if} \\
		\hspace{10mm}{\bf else} \\
		\hspace{15mm}Calculate the trial solution $\bm{u_{i}}$ according to (2). \\
		\hspace{15mm}Calculate the fitness of trial solution $\bm{u_{i}}$ and $FEs=FEs+1$. \\
		\hspace{10mm}{\bf end if} \\
		\hspace{10mm}{\bf if} the fitness of $u_{i}$ is better than that of $x_{i}$. \\
		\hspace{15mm}$\bm{x_{i}}=\bm{u_{i}}$. \\
		\hspace{15mm}{\bf if} the fitness of $\bm{x_{i}}$ is better than that of $\bm{x_{best}}$. \\
		\hspace{20mm}$\bm{x_{best}}=\bm{x_{i}}$. \\
		\hspace{15mm}{\bf end if} \\
		\hspace{10mm}{\bf end if} \\
		\hspace{5mm}{\bf end for} \\
		{\bf end while} \\
		Output the best solution $\bm{x_{best}}$ \\ \bottomrule
	\end{tabular}
\end{center}

\begin{figure}[h]
    \begin{flushleft}
    \begin{subfigure}{0.2\columnwidth}
        \begin{center}
        \includegraphics[width=\columnwidth, angle=270]{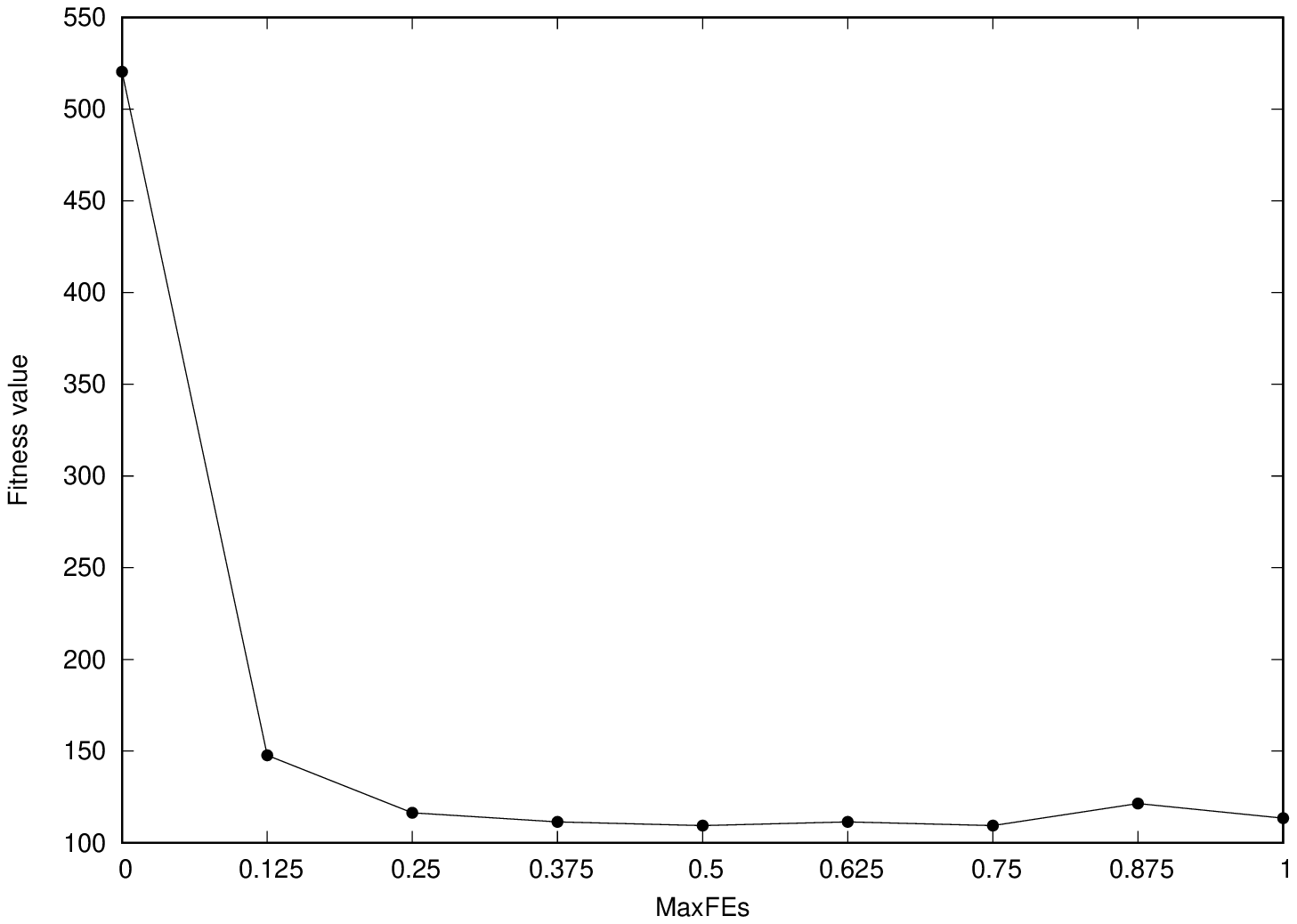}
        \caption{$F_{5}$}
        \end{center}
    \end{subfigure}
    \hspace{20mm}
    \begin{subfigure}{0.2\columnwidth}
        \begin{center}
        \includegraphics[width=\columnwidth, angle=270]{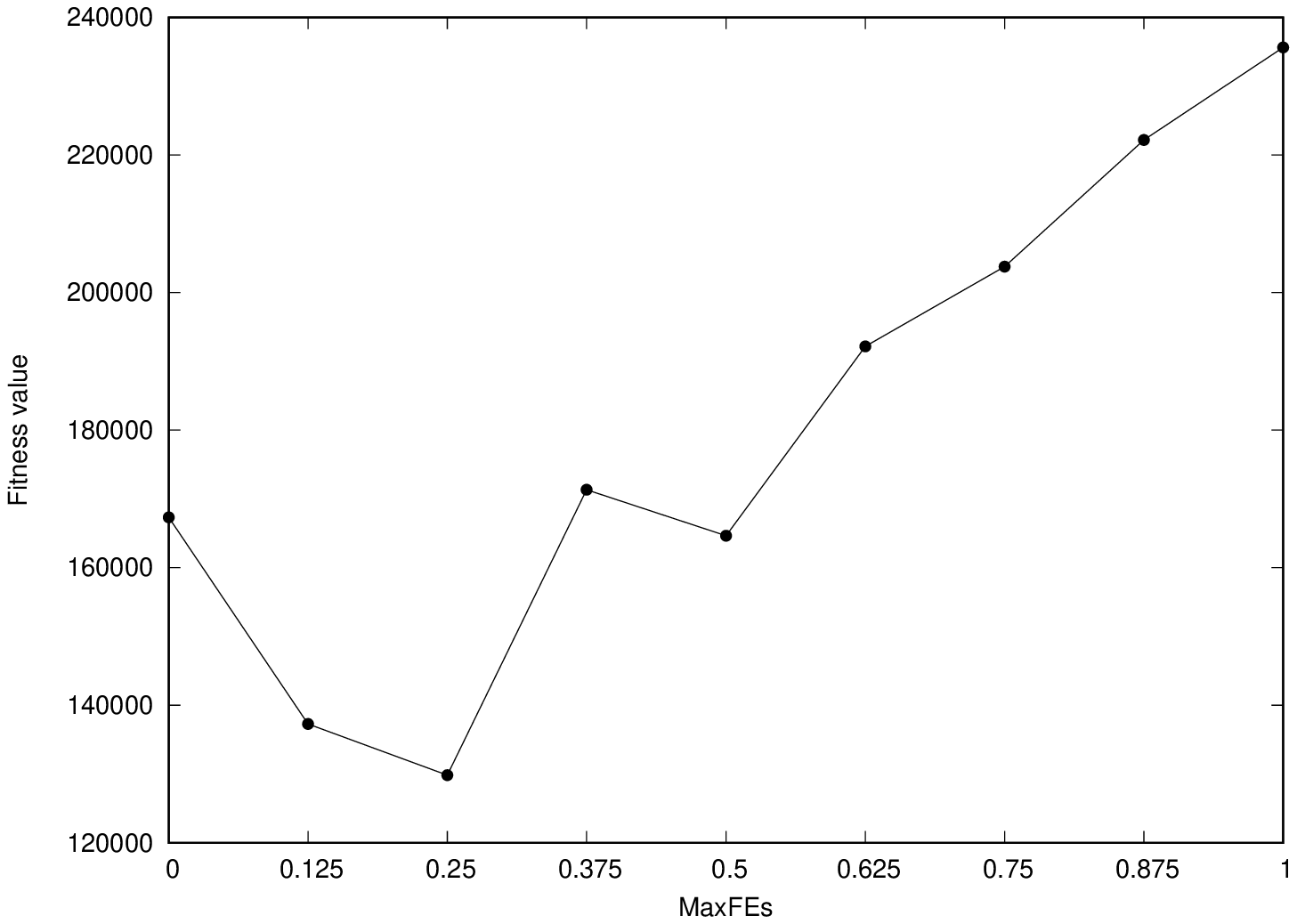}
        \caption{$F_{12}$}
        \end{center}
    \end{subfigure}
	\hspace{20mm}
    \begin{subfigure}{0.2\columnwidth}
        \begin{center}
        \includegraphics[width=\columnwidth, angle=270]{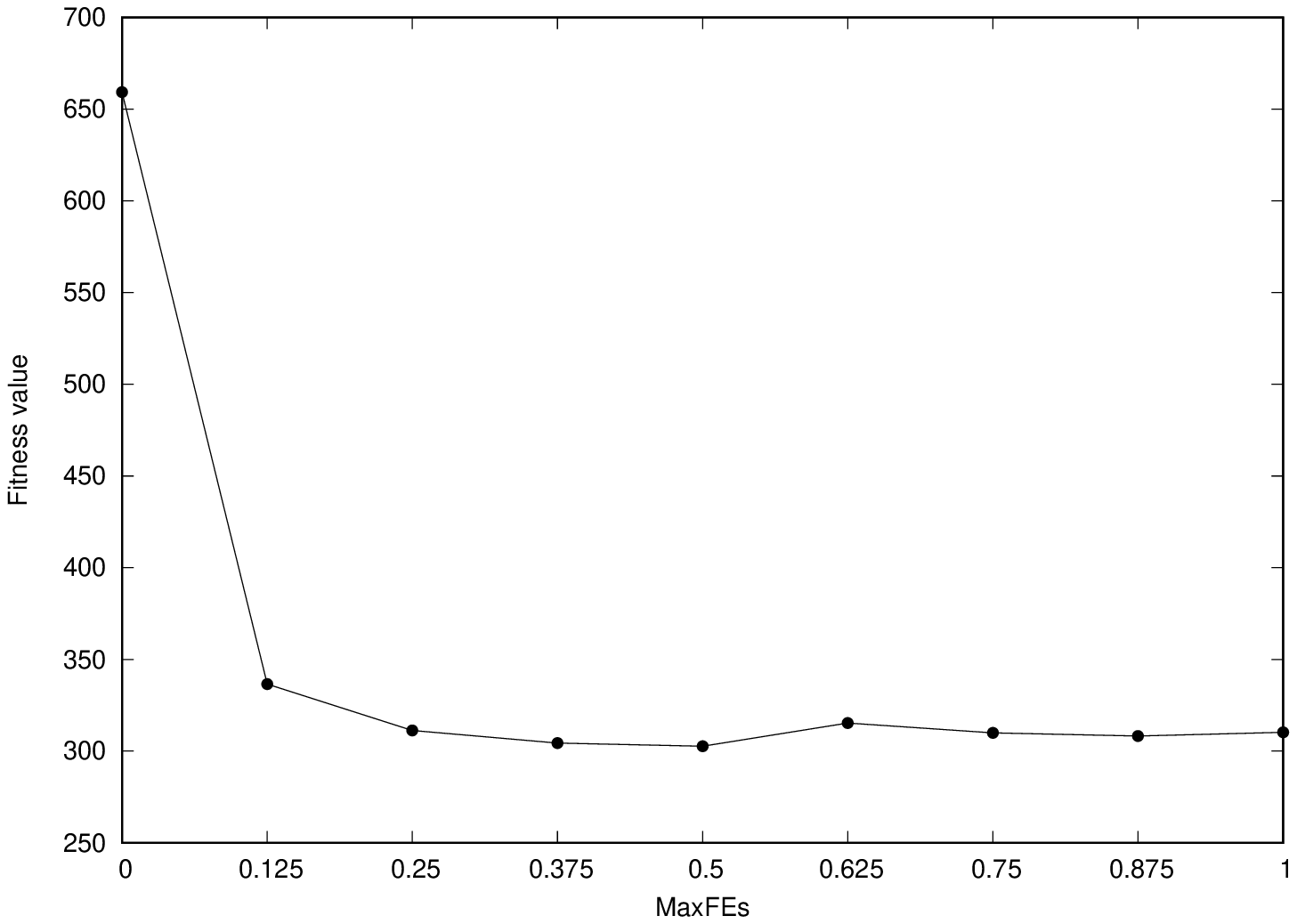}
        \caption{$F_{21}$}
        \end{center}
    \end{subfigure}
    
    \begin{subfigure}{0.2\columnwidth}
        \begin{center}
        \includegraphics[width=\columnwidth, angle=270]{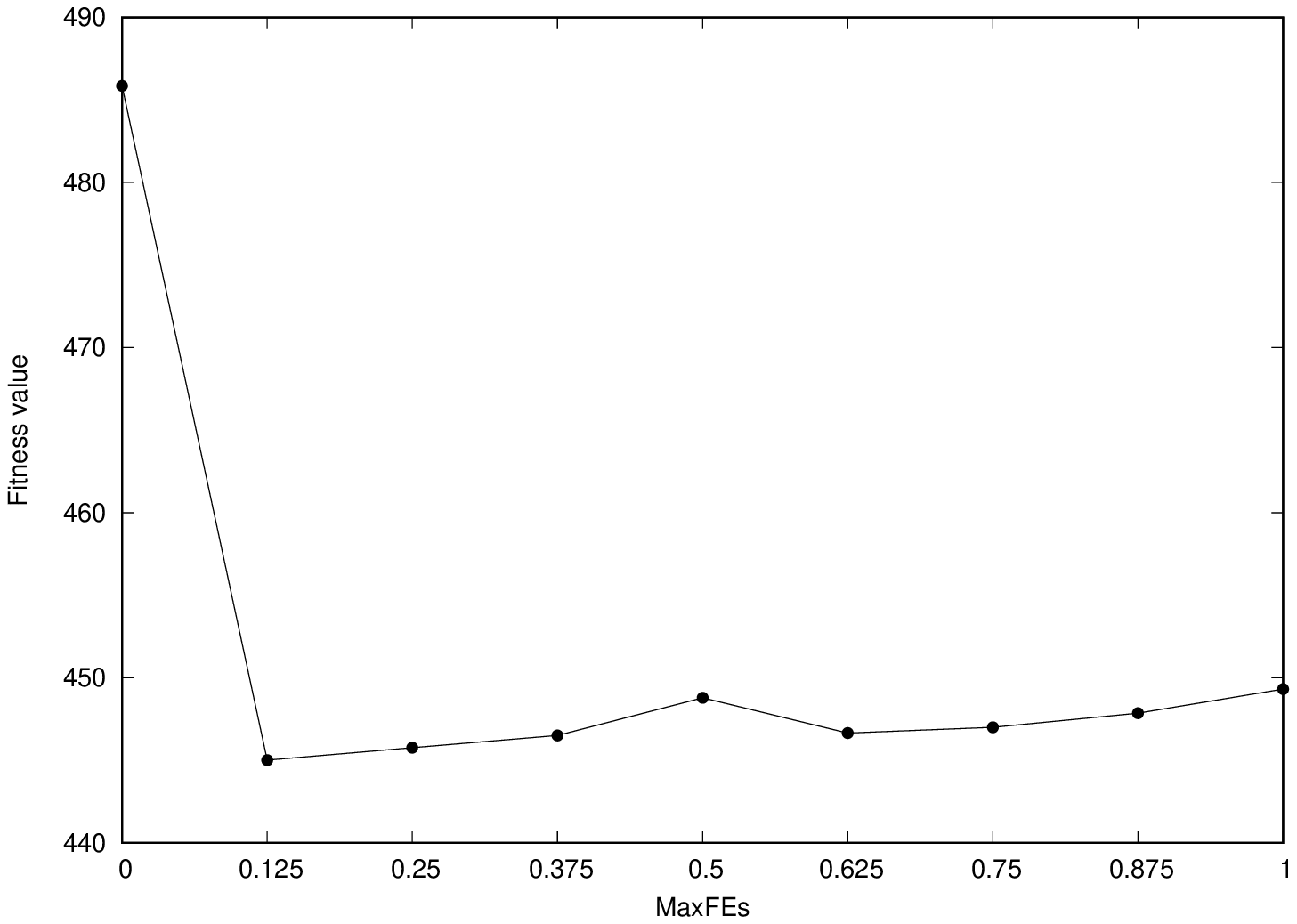}
        \caption{$F_{25}$}
        \end{center}
    \end{subfigure}
	\hspace{20mm}
    \begin{subfigure}{0.2\columnwidth}
        \begin{center}
        \includegraphics[width=\columnwidth, angle=270]{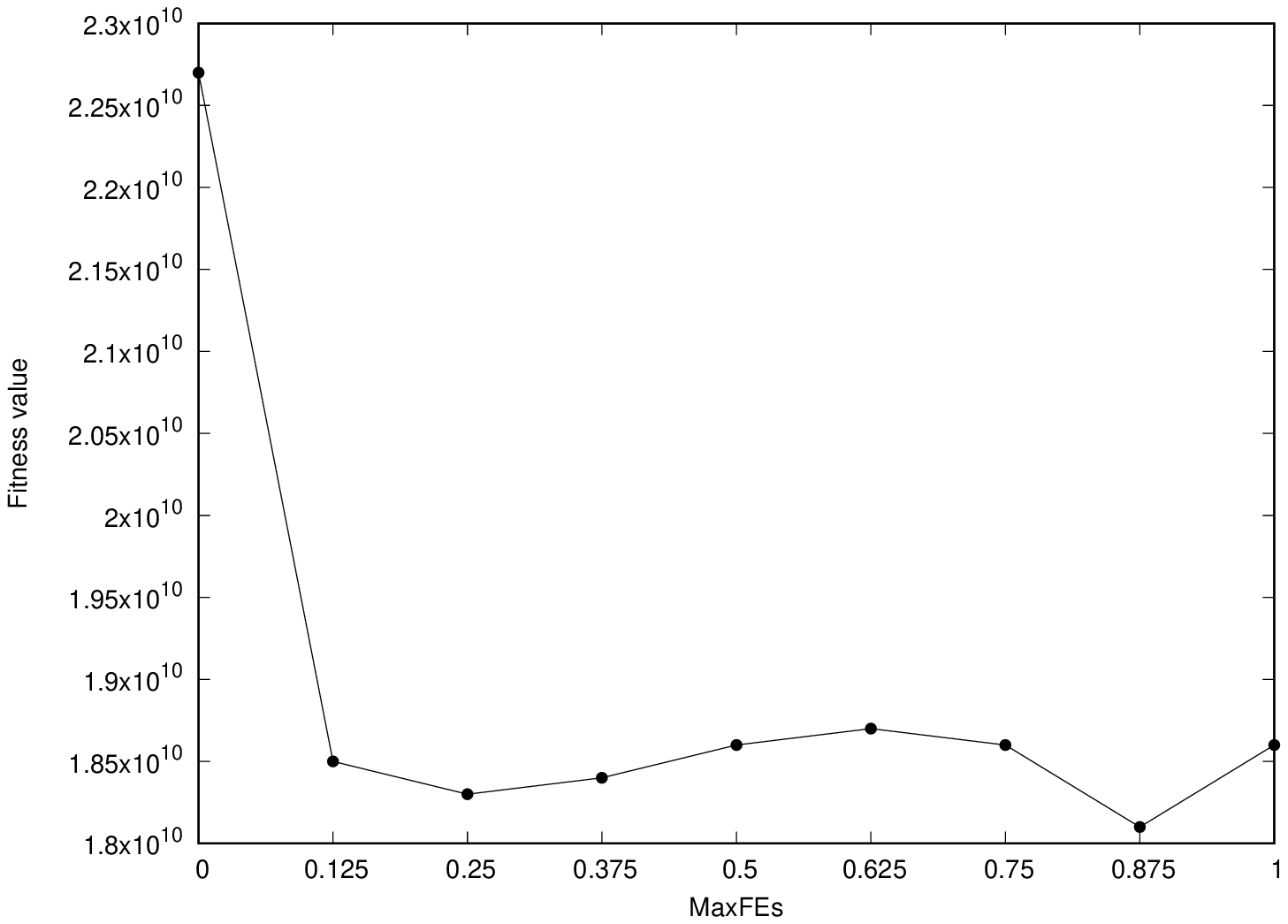}
        \caption{$F_{26}$}
        \end{center}
    \end{subfigure}
    \hspace{20mm}
    \begin{subfigure}{0.2\columnwidth}
        \begin{center}
        \includegraphics[width=\columnwidth, angle=270]{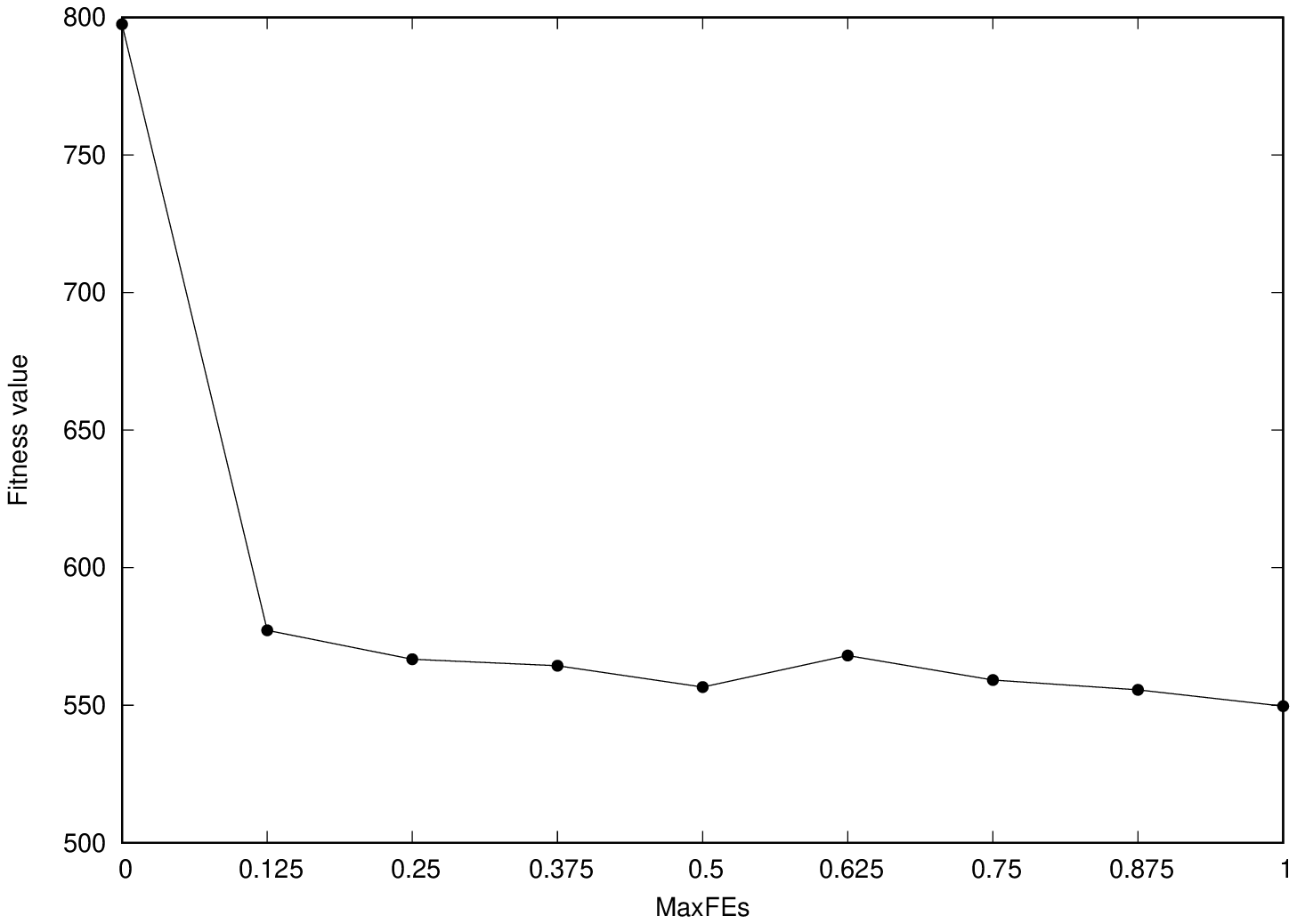}
        \caption{$F_{27}$}
        \end{center}
    \end{subfigure}
    \caption{Median fitness when the applicable time of the perturbation is changed in $1/8$ increments.}
    \end{flushleft}
\end{figure}%

\section{Numerical experiments}
We compare the proposed algorithm (DEP) to real-coded genetic algorithm (RGA), 
particle swarm optimization (PSO), differential evolution (DE), adaptive differential 
evolution with optional external archive (JADE), and success-story based parameter 
adaptation for differential evolution (SHADE) by CEC2017 
dataset. 
CEC2017 dataset consists of 29 functions ~\cite{REF CEC2017}. $F_{1}$ and $F_{2}$ are 
unimodal functions. $F_{3}$ to $F_{9}$ are multimodal functions. $F_{10}$ to 
$F_{19}$ are hybrid functions in which the variables are randomly divided into some 
subcomponents and then different basic functions are used for different subcomponents.
$F_{20}$ to $F_{29}$ are composition functions which merges the properties of the 
sub-functions better and maintains continuity around the global / local optima. 
Each algorithm searches a solution that minimizes the value of functions. 
We chose common parameters as follows: number of population $N=20$, 
number of dimension $D=10$ and $50$, and maximum number of function evaluation $MaxFEs=10000D$. 
We tried each algorithm 50 times and calculated the average. 
Parameters of 
each algorithm as follows: inertia weight $w=0.9-0.5FEs/MaxFEs$, both of weight parameters 
$\Phi_{1} = \Phi_{2} = 2.0$ for PSO, scale factor $F=0.8$, crossover rate $C_{r}=0.95$ 
for DE, parameter $c=0.2$, parameter $p=0.2$ for JADE and SHADE, and constant parameter $a=1.0$, constant parameter $R=200.0$ for DEP.

DEP generates a new solution using perturbation. The threshold is an important factor that determines how long the perturbation applied because 
the magnitude of perturbation decrease as the search progresses. Fig. 1. shows the median fitness when the adaptation time of the perturbation 
is changed in $1/8$ increments. The horizontal axis is MaxFEs and the vertical axis is the fitness value. The position of $0.0$ on the horizontal 
axis is the fitness of original DE. In the Fig. 1, the fitness takes almost the same value after $MaxFEs/4$, however for some functions, it tends 
to deteriorate gradually after $MaxFEs/4$. We adopt $MaxFEs/4$ as the adaptation time of perturbation. In this case, $R=200.0$ and $a=1.0$ are set.

Tables 1 and 3 are shown the average of best solution for each algorithm in 10 and 50 dimension, 
respectively. 
Tables 2 and 4 are shown the results of Wilcoxon signed rank test with a 
significance level of $5$ percent between DEP and other algorithm in 10 and 50 dimension ~\cite{REF Wilcoxson}. 
Three symbols ``+, -, =" indicate that DEP exhibits statistically 
better, worse than, and equal to the other algorithm, respectively. 
In Tables 1 and 2, DEP has the best solution for 18 out of the 29 functions. 
DEP obtains the statistically better solutions than RGA, PSO, DE, JADE, and SHADE on 29, 
21, 15, 15, and 15 cases. DEP shows the statistically worse solution than PSO, DE, JADE, 
SHADE on 3, 1, 7, and 7 cases. 
DEP displays the same performance than PSO, DE, JADE, and SHADE on 5, 13, 7, and 7 cases. 
In Tables 3 and 4, DEP performs the best solution for 14 out of the 29 functions. 
DEP gains the statistically better solutions than RGA, PSO, DE, JADE, and SHADE on 29, 25, 
14, 12, and 15 cases. DEP indicates the statistically worse solutions than PSO, JADE, and SHADE  
on  3, 7, and 6 cases. DEP represents the same performance than PSO, DE, JADE, and SHADE on 1, 15, 
10, and 8 cases. 
From these results, The performance of DEP is equal to or better than 
the existing algorithms. 

\section{Conclusions}
In this paper, we have presented differential evolution with 
perturbation (DEP). In order to show the effectiveness 
of the proposed algorithm, we compared DEP with RGA, PSO, DE, JADE, and SHADE using CEC2017 
dataset. The performance of DEP was equal to or better than 
the existing algorithms. For future studies, we conduct 
theoretical considerations of the proposed algorithm.

\begin{table}[h]
	\caption{Average of best solutions for each algorithm in 10 dimensional functions.}
	\label{tab:1}
	\begin{center}
		\begin{tabular}{p{15mm} p{21mm} p{21mm} p{21mm} p{21mm} p{21mm} p{21mm}} \toprule
			Function  &        RGA         &         PSO        &         DE         &        JADE        &         SHADE      &        DEP              \\ \midrule
			 $F_{1}$  & $9.87\times10^{9}$ & $2.41\times10^{7}$ & $1.11\times10^{2}$ & $1.01\times10^{3}$ & $2.38\times10^{-5}$&    \bm{$0.0$}           \\
			 $F_{2}$  & $2.99\times10^{4}$ &     \bm{$0.0$}     &     \bm{$0.0$}     &     \bm{$0.0$}     &     \bm{$0.0$}     &    \bm{$0.0$}           \\
			 $F_{3}$  & $9.40\times10^{2}$ & $2.23\times10^{1}$ & $3.88\times10^{-1}$&       $1.95$       & $8.94\times10^{-2}$&    \bm{$0.0$}           \\
			 $F_{4}$  & $1.18\times10^{4}$ & $9.39\times10^{1}$ & $1.05\times10^{2}$ & $5.52\times10^{1}$ & $1.77\times10^{1}$ & \bm{$1.35\times10^{1}$} \\
			 $F_{5}$  & $5.64\times10^{-3}$& $1.59\times10^{-5}$& $2.07\times10^{-4}$& $2.22\times10^{-8}$& \bm{$1.14\times10^{-8}$}& $9.11\times10^{-5}$\\
			 $F_{6}$  & $1.47\times10^{4}$ & $2.36\times10^{1}$ & $4.25\times10^{1}$ & $9.16\times10^{1}$ & \bm{$2.18\times10^{1}$} & $3.07\times10^{1}$ \\
			 $F_{7}$  & $4.16\times10^{2}$ & $3.08\times10^{1}$ & $5.60\times10^{1}$ & $3.77\times10^{1}$ & \bm{$2.52\times10^{1}$} & $5.55\times10^{1}$ \\
			 $F_{8}$  & $9.13\times10^{-1}$& \bm{$6.36\times10^{-1}$}& \bm{$6.36\times10^{-1}$}& \bm{$6.36\times10^{-1}$}& \bm{$6.36\times10^{-1}$}& \bm{$6.36\times10^{-1}$}\\
			 $F_{9}$  & $2.76\times10^{3}$ & $4.01\times10^{2}$ & $3.59\times10^{2}$ & $8.04\times10^{1}$ & \bm{$5.80\times10^{1}$} & $2.59\times10^{2}$ \\
			 $F_{10}$ & $1.08\times10^{8}$ & $1.55\times10^{3}$ & $2.02\times10^{3}$ & $1.21\times10^{1}$ & $4.16\times10^{1}$ &       \bm{$2.71$}       \\
			 $F_{11}$ & $2.45\times10^{8}$ & $1.77\times10^{5}$ & $4.22\times10^{6}$ & $2.96\times10^{4}$ & $3.99\times10^{3}$ & \bm{$1.42\times10^{2}$} \\
			 $F_{12}$ & $3.34\times10^{8}$ & $8.40\times10^{3}$ & $3.35\times10^{4}$ & $1.86\times10^{3}$ & $1.23\times10^{2}$ & \bm{$2.38\times10^{1}$} \\
			 $F_{13}$ & $3.70\times10^{4}$ & $3.34\times10^{1}$ & $1.55\times10^{1}$ & $2.04\times10^{1}$ & $2.62\times10^{1}$ & \bm{$1.09\times10^{1}$} \\
			 $F_{14}$ & $2.06\times10^{8}$ & $4.55\times10^{3}$ & $3.62\times10^{4}$ & $5.30\times10^{1}$ & $1.94\times10^{1}$ &       \bm{$1.78$}       \\
			 $F_{15}$ & $8.30\times10^{3}$ & $3.34\times10^{2}$ & $4.04\times10^{2}$ & $2.58\times10^{1}$ & $3.12\times10^{2}$ & \bm{$1.40\times10^{2}$} \\
			 $F_{16}$ & $4.72\times10^{7}$ & $3.13\times10^{1}$ & $2.21\times10^{1}$ &       \bm{$9.58$}       &       $9.63$       & $1.21\times10^{1}$ \\
			 $F_{17}$ & $4.66\times10^{7}$ & $1.56\times10^{4}$ & $1.80\times10^{2}$ &       \bm{$8.60$}       & $2.90\times10^{1}$ & $1.13\times10^{1}$ \\
			 $F_{18}$ & $1.73\times10^{10}$& $1.35\times10^{3}$ & $1.52\times10^{3}$ &       $1.73$       &       $2.37$       &       \bm{$1.32$}       \\
			 $F_{19}$ & $1.59\times10^{3}$ & $5.80\times10^{1}$ & $1.95\times10^{1}$ &       $5.96$       &       \bm{$4.04$}       & $2.02\times10^{1}$ \\
			 $F_{20}$ & $1.22\times10^{3}$ & $2.13\times10^{2}$ & $1.89\times10^{2}$ & $1.10\times10^{2}$ & $1.30\times10^{2}$ & \bm{$1.02\times10^{2}$} \\
			 $F_{21}$ & $1.64\times10^{2}$ & $1.04\times10^{2}$ & $1.04\times10^{2}$ & $1.02\times10^{2}$ & $1.02\times10^{2}$ & \bm{$1.01\times10^{2}$} \\
			 $F_{22}$ & $2.39\times10^{3}$ & $2.08\times10^{2}$ & $2.18\times10^{2}$ & $1.28\times10^{2}$ & $1.49\times10^{2}$ & \bm{$1.26\times10^{2}$} \\
			 $F_{23}$ & $3.36\times10^{3}$ & $2.93\times10^{2}$ & $4.90\times10^{2}$ & \bm{$1.23\times10^{2}$} & $1.92\times10^{2}$ & $2.00\times10^{2}$ \\
			 $F_{24}$ & $2.92\times10^{9}$ & $9.39\times10^{4}$ & $4.01\times10^{2}$ & $4.66\times10^{2}$ & $4.08\times10^{2}$ & \bm{$4.00\times10^{2}$} \\
			 $F_{25}$ & $9.27\times10^{8}$ & $6.29\times10^{8}$ & $9.83\times10^{8}$ & \bm{$5.65\times10^{7}$} & $6.25\times10^{8}$ & $9.73\times10^{8}$ \\
			 $F_{26}$ & $4.06\times10^{2}$ & $4.28\times10^{2}$ & $4.14\times10^{2}$ & \bm{$3.74\times10^{2}$} & $3.98\times10^{2}$ & $3.97\times10^{2}$ \\
			 $F_{27}$ & $1.39\times10^{8}$ & $1.88\times10^{5}$ & $2.23\times10^{5}$ & $3.33\times10^{3}$ & $1.31\times10^{3}$ & \bm{$7.46\times10^{2}$} \\
			 $F_{28}$ & $7.48\times10^{4}$ & $2.92\times10^{3}$ & $1.45\times10^{3}$ & $1.30\times10^{3}$ & $1.58\times10^{3}$ & \bm{$7.54\times10^{2}$} \\
			 $F_{29}$ & $1.57\times10^{8}$ & $1.58\times10^{6}$ & $1.24\times10^{6}$ & \bm{$2.60\times10^{5}$} & $1.42\times10^{6}$ & $1.07\times10^{6}$ \\ \bottomrule
		\end{tabular}
	\end{center}
\end{table}

\begin{table}[h]
	\caption{Results of Wilcoxon signed rank sum test between DEP and other algorithm in 10 dimension.}
	\label{tab:2}
	\begin{center}
		\begin{tabular}{p{41mm}p{34mm}p{34mm}p{34mm}} \\ \toprule
			Algorithms &        +        &          =          &          -        \\ \midrule
			   RGA     &       $29$      &         $0$         &          $0$      \\
			   PSO     &       $21$      &         $5$         &          $3$      \\
			    DE     &       $15$      &         $13$        &         $1$      \\
				JADE   &       $15$      &         $7$         &         $7$      \\
				SHADE  &       $15$      &         $7$         &         $7$      \\ \bottomrule
		\end{tabular}
	\end{center}
\end{table}

\begin{table}[h]
	\caption{Average of best solutions for each algorithm in 50 dimensional functions.}
	\label{tab:3}
	\begin{center}
		\begin{tabular}{p{15mm} p{21mm} p{21mm} p{21mm} p{21mm} p{21mm} p{21mm}} \toprule
			Function  &        RGA         &         PSO        &         DE         &        JADE        &         SHADE      &        DEP              \\ \midrule
			 $F_{1}$  & $2.61\times10^{11}$& $1.78\times10^{9}$ & $6.15\times10^{3}$ & \bm{$2.44\times10^{3}$} & $3.19\times10^{3}$ & $9.07\times10^{3}$   \\
			 $F_{2}$  & $4.17\times10^{5}$ & $1.53\times10^{3}$ &     \bm{$0.0$}     &$8.56\times10^{-10}$&     \bm{$0.0$}     &    \bm{$0.0$}           \\
			 $F_{3}$  & $8.99\times10^{4}$ & $2.89\times10^{2}$ & \bm{$7.11\times10^{1}$} & $7.65\times10^{1}$ & $1.00\times10^{2}$ & $8.07\times10^{1}$    \\
			 $F_{4}$  & $2.84\times10^{5}$ & $3.88\times10^{3}$ & $6.07\times10^{2}$ & $9.66\times10^{2}$ & $2.18\times10^{3}$ & \bm{$1.19\times10^{2}$} \\
			 $F_{5}$  & $1.34\times10^{-1}$& $5.22\times10^{-4}$& $4.57\times10^{-3}$& $6.35\times10^{-5}$& \bm{$3.17\times10^{-5}$}& $1.35\times10^{-3}$\\
			 $F_{6}$  & $3.80\times10^{5}$ & $3.79\times10^{3}$ & $8.41\times10^{2}$ & $1.09\times10^{3}$ & $2.53\times10^{3}$ & \bm{$2.28\times10^{2}$} \\
			 $F_{7}$  & $1.27\times10^{4}$ & \bm{$3.71\times10^{2}$} & $5.39\times10^{2}$ & $9.88\times10^{2}$ & $4.78\times10^{2}$ & $5.34\times10^{2}$ \\
			 $F_{8}$  &       $4.97$       &       $4.73$       &       \bm{$4.70$}       &       $4.71$       &       $4.71$       &       $4.71$       \\
			 $F_{9}$  & $3.61\times10^{4}$ & $3.93\times10^{3}$ & $5.50\times10^{3}$ & \bm{$1.08\times10^{3}$} & $4.33\times10^{3}$ & $5.84\times10^{3}$ \\
			 $F_{10}$ & $2.74\times10^{10}$& $4.27\times10^{6}$ & $1.34\times10^{5}$ & $9.51\times10^{2}$ & $4.98\times10^{3}$ & \bm{$5.17\times10^{2}$} \\
			 $F_{11}$ & $1.01\times10^{11}$& $1.42\times10^{9}$ & $1.66\times10^{7}$ & $1.93\times10^{6}$ & $6.47\times10^{5}$ & \bm{$1.45\times10^{5}$} \\
			 $F_{12}$ & $1.69\times10^{11}$& $9.36\times10^{8}$ & $4.72\times10^{7}$ & \bm{$1.00\times10^{4}$} & $1.79\times10^{4}$ & $1.16\times10^{4}$ \\
			 $F_{13}$ & $3.77\times10^{7}$ & $5.24\times10^{5}$ & $2.39\times10^{4}$ & $5.35\times10^{5}$ & $5.17\times10^{4}$ & \bm{$1.01\times10^{4}$} \\
			 $F_{14}$ & $1.12\times10^{11}$& $5.17\times10^{6}$ & $1.52\times10^{4}$ & $9.55\times10^{3}$ & $4.73\times10^{4}$ & \bm{$9.09\times10^{3}$} \\
			 $F_{15}$ & $2.11\times10^{10}$& $1.40\times10^{5}$ & $1.17\times10^{3}$ & \bm{$2.14\times10^{2}$} & $9.83\times10^{2}$ & $6.89\times10^{2}$ \\
			 $F_{16}$ & $6.28\times10^{16}$& $1.84\times10^{4}$ & $2.20\times10^{3}$ & $6.66\times10^{2}$ & $9.88\times10^{2}$ & \bm{$6.01\times10^{2}$} \\
			 $F_{17}$ & $1.11\times10^{8}$ & $4.76\times10^{5}$ & \bm{$1.25\times10^{3}$} & $9.57\times10^{5}$ & $2.95\times10^{5}$ & $4.63\times10^{4}$ \\
			 $F_{18}$ & $3.71\times10^{15}$& $3.16\times10^{9}$ & $7.42\times10^{8}$ & \bm{$1.40\times10^{4}$} & $2.08\times10^{4}$ & $1.78\times10^{4}$ \\
			 $F_{19}$ & $4.50\times10^{4}$ & $6.23\times10^{2}$ & $2.93\times10^{2}$ & $2.58\times10^{2}$ & $2.68\times10^{2}$ & \bm{$2.35\times10^{2}$} \\
			 $F_{20}$ & $1.97\times10^{4}$ & $3.79\times10^{3}$ & $7.60\times10^{2}$ & $5.41\times10^{2}$ & $1.65\times10^{3}$ & \bm{$2.60\times10^{2}$} \\
			 $F_{21}$ & $4.25\times10^{3}$ & $7.86\times10^{2}$ & $4.19\times10^{2}$ & \bm{$1.22\times10^{2}$} & $1.85\times10^{2}$ & $4.83\times10^{2}$ \\
			 $F_{22}$ & $8.88\times10^{4}$ & $1.01\times10^{4}$ & $1.47\times10^{3}$ & $6.36\times10^{2}$ & $1.19\times10^{3}$ & \bm{$2.54\times10^{2}$} \\
			 $F_{23}$ & $5.19\times10^{4}$ & $6.30\times10^{3}$ & $5.72\times10^{2}$ & $9.85\times10^{2}$ & $8.11\times10^{2}$ & \bm{$2.28\times10^{2}$} \\
			 $F_{24}$ & $2.75\times10^{11}$& $1.16\times10^{8}$ & $5.16\times10^{2}$ & \bm{$4.61\times10^{2}$} & $5.20\times10^{2}$ & $4.96\times10^{2}$ \\
			 $F_{25}$ & $3.69\times10^{10}$& $2.38\times10^{10}$& $3.05\times10^{10}$& \bm{$9.96\times10^{9}$} & $1.66\times10^{10}$& $1.85\times10^{10}$ \\
			 $F_{26}$ & $1.33\times10^{3}$ & $1.09\times10^{3}$ & $8.77\times10^{2}$ & \bm{$5.41\times10^{2}$} & $5.85\times10^{2}$ & $5.73\times10^{2}$ \\
			 $F_{27}$ & $4.02\times10^{10}$& $1.02\times10^{8}$ & $7.55\times10^{3}$ & \bm{$2.63\times10^{3}$} & $1.05\times10^{5}$ & $2.77\times10^{3}$ \\
			 $F_{28}$ & $2.36\times10^{14}$& $1.11\times10^{8}$ & $4.32\times10^{8}$ & $1.17\times10^{5}$ & $2.79\times10^{5}$ & \bm{$6.32\times10^{4}$} \\
			 $F_{29}$ & $1.90\times10^{16}$& $5.08\times10^{8}$ & $1.64\times10^{8}$ & $1.98\times10^{6}$ & $3.41\times10^{6}$ & \bm{$1.56\times10^{6}$} \\ \bottomrule
		\end{tabular}
	\end{center}
\end{table}

\begin{table}[h]
	\caption{Results of Wilcoxon signed rank sum test between DEP and other algorithm in 50 dimension.}
	\label{tab:4}
	\begin{center}
		\begin{tabular}{p{41mm}p{34mm}p{34mm}p{34mm}} \\ \toprule
			Algorithms &        +        &          =          &          -        \\ \midrule
			   RGA     &       $29$      &         $0$         &          $0$      \\
			   PSO     &       $25$      &         $1$         &          $3$      \\
			    DE     &       $14$      &         $15$         &         $0$      \\
				JADE     &       $12$      &         $10$         &         $7$      \\
				SHADE     &       $15$      &         $8$         &         $6$      \\ \bottomrule
		\end{tabular}
	\end{center}
\end{table}

\end{document}